\documentclass[10pt,letterpaper]{article}

\usepackage{iccv}
\usepackage{times}
\usepackage{epsfig}
\usepackage{graphicx}
\usepackage{amsmath}
\usepackage{amssymb}
\usepackage{subfigure}
\usepackage{multirow}

\usepackage[breaklinks=true,bookmarks=false]{hyperref}

\iccvfinalcopy % *** Uncomment this line for the final submission

\def\iccvPaperID{****} % *** Enter the ICCV Paper ID here

\begin{document}

%%%%%%%%% TITLE
\title{EventNet Version 1.1 Technical Report}

\author{Dongang Wang, Zheng Shou, Hongyi Liu, and Shih-Fu Chang\\
Columbia University\\
New York, NY, USA\\
{\tt\small \{dw2648, zs2262, hl2906, sc250\}@columbia.edu}
}

\maketitle
%\thispagestyle{empty}

%%%%%%%%% ABSTRACT
\begin{abstract}

EventNet is a large-scale video corpus and event ontology consisting of 500 events associated with event-specific concepts. In order to improve the quality of the current EventNet, we conduct the following steps and introduce \textbf{EventNet version 1.1}: (1) manually verify the correctness of event labels for all videos; (2) remove the YouTube user bias by limiting the maximum number of videos in each event from the same YouTube user as 3; (3) remove the videos which are currently not accessible online; (4) remove the video belonging to multiple event categories. After the above procedure, some events may contain only a small number of videos, and therefore we crawl more videos for those events to ensure every event will contain more than 50 videos. Finally, EventNet version 1.1 contains 67,641 videos, 500 events, and 5,028 event-specific concepts. In addition, we train a Convolutional Neural Network (CNN) model for event classification via fine-tuning AlexNet using EventNet version 1.1. Then we use the trained CNN model to extract $\tt FC7$ layer feature and train binary classifiers using linear SVM for each event-specific concept. We believe this new version of EventNet will significantly facilitate research in computer vision and multimedia, and we will put it online for the public to download in the future.

\end{abstract}

\tableofcontents

\newpage

\section{Introduction}

EventNet is a large-scale video corpus and event ontology associated with event-specific concepts. Currently, EventNet ontology consists of 500 events, 4,490 event-specific concepts, and 99,382 videos \cite{eventnet}. However, the quality of video corpus still has much room for improvement in order to make EventNet more useful to the computer vision and multimedia researchers. In this report, section 2 refines the video corpus from several perspectives, and section 3 trains event and concept classification model. Eventually, EventNet version 1.1 contains 67,641 videos, 500 events, and 5,028 event-specific concepts.

\section{Video corpus refinement}

\subsection{Video removal}\label{removal}

\noindent\textbf{Remove videos of incorrect event label}: Using the approach stated in \cite{eventnet}, some videos can have incorrect event labels. For example, for the event ``running'', if the title of a video on YouTube contains ``running'', then it may be downloaded and added into EventNet. However, there is a TV show called ``Running Man'' which is irrelevant with the event, and therefore requires manual checking and removal. We invited around 10 annotators for manual verification. Each event is assigned to at least one annotator and one annotator is asked to randomly check the annotation results. Finally, there are in total 10,521 videos removed by manual verification. Without label noises, researchers can train more accurate event detection model and achieve better performances in other tasks such as zero-shot retrieval.

\hfill\break\noindent\textbf{Remove YouTube user bias}: When Ye \textit{et al.} \cite{eventnet} previously crawled videos, they did not consider the video uploader's information. However, if many videos for one event are from the same YouTube user, the classifier trained for this event would actually capture the characteristics of this YouTube user rather than the event. Therefore it is important to remove the YouTube user bias. In each event, if the number of videos from the same YouTube user is more than 3, we randomly select three of them and remove the rest videos. Consequently, 17,257 videos are removed during the YouTube user bias removal. For some events, previously almost all the videos were downloaded from one single playlist on YouTube, and therefore now most videos are removed, for example ``barbecue'' only contains 9 videos after removing biased videos. In addition, removing YouTube user bias can also help facilitate building personalized event ontology, because the video corpus becomes diverse instead of being biased to some YouTube users.

\hfill\break\noindent\textbf{Remove inaccessible videos}: Till now, some videos in EventNet video corpus become inaccessible on YouTube. In EventNet version 1.1, we remove these inaccessible videos, which count for 6,130 out of the 99,382 videos in the original EventNet video corpus. However, videos are being generated and vanished every day, and there are many videos inaccessible after our attempt. We have stored all the videos and corresponding information (uploader, uploaded time, title, etc.) on our server for later checking.

\hfill\break\noindent\textbf{Remove repeated videos}: Among the original 99,382 videos, the number of unique videos is 97,576, which means that some videos appear in different event categories. For example, there are some repeated video between ``clean silverware'' and ``clean jewelry''. In \cite{eventnet}, the identical video belonging to different events was considered as different videos. Here we identify and then remove these repeated videos from the corpus. We use the rest videos for later experiments. However, we still have another version that keep the repeated videos for later experiments.

\subsection{Video corpus augmentation}

After the above video removal, we end up with 14 events which contain less than 50 videos, which are ``ballroom dance'', ``barbecue'', ``breed ducks'', ``combat sports'', ``cook poultry'', ``do yoyo tricks'', ``field hockey'', ``organize a bookshelf'', ``pack a moving truck'', ``remove facial warts'', ``running'', ``snowboarding'', ``visit theme park'', and ``welding''. For all these 14 events, we crawl new videos from YouTube and also make sure the criteria mentioned in section \ref{removal} are still satisfied.

\begin{figure}[h]
\centering
\includegraphics[width=\textwidth]{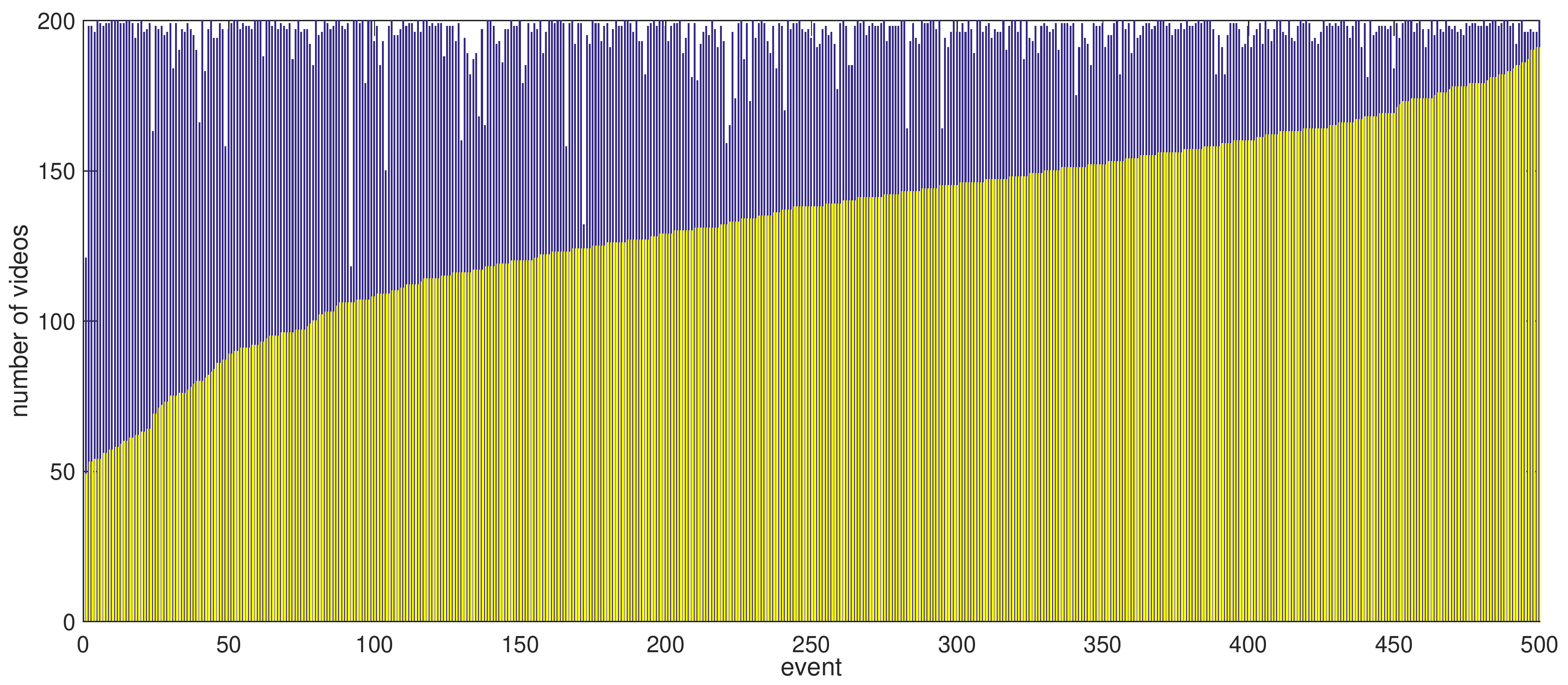}
\caption{The number of videos for each event, sorted by the number of videos in EventNet version 1.1. The yellow bars stand for the number of videos in EventNet version 1.1, and the blue bars stand for the original EventNet video corpus.}
\label{STATs}
\end{figure}

\subsection{Statistics}

Finally, the total number of videos in EventNet version 1.1 is 67,641. Figure \ref{STATs} illustrates the number of videos for each event in the original EventNet video corpus and in this version 1.1. The total number of events is still 500, and each event contain around 10 concepts. Each event has more than 50 videos. The total number of event-specific concepts is 5,028.

\section{Event and event-specific concept classification model}

Following \cite{eventnet}, we train event and event-specific concept classification model for EventNet version 1.1.

\subsection{Event model}

For each event, we randomly select 70\% videos as training set, 10\% videos as validation set, and 20\% videos as test set. Also, some videos belonging to the same event may come from the YouTube same user, and therefore we keep the videos uploaded by the same user together in one of these three sets.

We train Convolutional Neural Network (CNN) model for event classification. For each video, we uniformly extract around 40 frames and treat them as key frames. We adopt AlexNet \cite{alex}, which is pre-trained on ImageNet \cite{deng2009imagenet}, and replace the $\tt fc8$ layer by 500 event nodes. Different from the previous version, we add a background class and randomly pick some videos as training data from categories in ActivityNet  \cite{caba2015activitynet} without semantic overlap with our 500 event  categories. Then we fine-tune this model on the training set in EventNet version 1.1. We set the total number of iterations is 450,000, the batch size as 256, the momentum as 0.9, and the weight decay as 0.0005. We set the learning rate as 0.01 and divide it by 10 for every 10000 iterations. We use Titan X GPU of 12G memory during training.

In some scenarios, researchers may be interested in binary event classifier instead of the multi-class CNN classification model. So we also train linear SVM binary classifiers. For each event, in the training set, we regard frames of all videos belonging to this event as positive and randomly sample 5K frames from videos of other events as negative. Later on, we also train another version of SVM classifiers using 20K negative samples per event. For each frame, we extract 4096-dim $\tt fc7$ layer feature and do L2 normalization. Finally, we use LIBSVM \cite{libsvm} with C=100 to train binary classifier for each event. Note that these 500 binary classifiers can also be used for doing multi-class classification. We treat the confidence scores from 500 classifiers as the estimated probabilities for being the corresponding events.

\begin{table}[h]
\centering
\caption{Accuracies on the test set in EventNet version 1.1.}
\label{my-label}
\begin{tabular}{|c|c|c|}
\hline
                                               & multi-class   & binary                  \\ \hline
\multirow{2}{*}{SVM (5K negative samples)}  & Top-1: 0.1842 & \multirow{2}{*}{0.6342} \\ \cline{2-2}
                                               & Top-5: 0.3809 &                         \\ \hline
\multirow{2}{*}{SVM (20K negative samples)}  & Top-1: 0.1933 & \multirow{2}{*}{0.7422} \\ \cline{2-2}
 & Top-5: 0.3799 &                         \\ \hline
\multirow{2}{*}{CNN}                           & Top-1: 0.3067 & \multirow{2}{*}{-}      \\ \cline{2-2}
                                               & Top-5: 0.5327 &                         \\ \hline
\end{tabular}
\label{res}
\end{table}

Table \ref{res} reports accuracies on the test set in EventNet version 1.1 for the above event classification models. When evaluating multi-class classification performance, we directly use the test set in EventNet version 1.1. We first apply classification model on every frame and then do max pooling over time to get a 500-dim vector for each video. CNN model achieves better result than SVM, because when the number of negative training samples for training SVM is relatively small given the large scale of EventNet. As for binary classification, only SVM classifiers can suit for this task. For each event, the videos in EventNet version 1.1 test set are regarded as positive and we randomly sample the equal number of videos from other event categories as negative test samples.

\subsection{Concept model}

Based on the new video corpus in EventNet version 1.1, we follow the approach in \cite{eventnet} to update event-specific concepts. In addition, we remove the concepts that only occur in less than 3 videos, and manually verify correctness of all event-specific concepts. Finally, the total number of event-specific concepts is 5,028 and the number of unique concepts is 1,512.

Similarly as training SVM classifier for event, we train linear SVM binary classification model for event-specific concept. For each event-specific concept, all frames from the videos associated with this event-specific concept are treated as positive, and we randomly sample the same number of videos from the other events and regard their frames as negative training samples. For each frame, we extract 4096-dim $\tt fc7$ layer feature and do L2 normalization. Finally, we use LIBSVM \cite{libsvm} with C=100 to train binary classifier for each event-specific concept.

\section{Conclusions}

In this report, we improve the quality of the current EventNet via manual verification of event labels for all videos, YouTube user bias removal, inaccessible videos removal, and filtering out the video belonging to multiple event categories. Finally, EventNet version 1.1 contains 67,641 videos, 500 events, and 5,028 event-specific concepts. Furthermore, we crawl more videos for those events to ensure every event will contain more than 50 videos. In addition, we also train a Convolutional Neural Network (CNN) model for event classification and also train linear SVM classifiers for event-specific concepts based on $\tt FC7$ layer feature. Given the higher quality, EventNet version 1.1 will significantly facilitate research in computer vision and multimedia.

\section{Acknowledgment}

This work is supported by the Intelligence Advanced Research Projects Activity (IARPA) via Department of Interior National Business Center contract number D11PC20071. The U.S. Government is authorized to reproduce and distribute reprints for Governmental purposes notwithstanding any copyright annotation thereon. Disclaimer: The views and conclusions contained herein are those of the authors and should not be interpreted as necessarily representing the official policies or endorsements, either expressed or implied, of IARPA, DOI-NBC, or the U.S. Government. We appreciate Fudan University for the help with manual verification of event labels for all videos.

{\small
\bibliographystyle{ieee}
\bibliography{egbib}
}

\end{document}